\documentclass[runningheads]{llncs}
\usepackage[T1]{fontenc}
\usepackage{algorithm}
\usepackage{algpseudocode}
\usepackage{multirow}
\usepackage{amsmath}
\usepackage{amssymb}
\usepackage{float}
\usepackage{graphicx}
\usepackage{caption}
\usepackage{subcaption}
\usepackage{booktabs}
\usepackage{xcolor}
\usepackage{multirow}
\usepackage[normalem]{ulem}
\useunder{\uline}{\ul}{}
\usepackage{algorithm}
\usepackage{algpseudocode}
\usepackage{url}

\begin{document}

\title{TREX: Trajectory Explanations for Multi-Objective Reinforcement Learning}
%
%
\author{Dilina Rajapakse\inst{1}\orcidID{0000-0001-9722-9108} \and
Juan C. Rosero\inst{1}\orcidID{0009-0009-7848-3966} \and
Ivana Dusparic\inst{1}\orcidID{0000-0003-0621-5400}}
\authorrunning{Rajapakse et al.}
%
\institute{Trinity College Dublin, Ireland\\ \email{\{rajapakd,roserolj,ivana.dusparic\}@tcd.ie}}
\maketitle             

\begin{abstract}

    Reinforcement Learning (RL) has demonstrated its ability to solve complex decision-making problems in a variety of domains, by optimizing reward signals obtained through interaction with an environment. However, many real-world scenarios involve multiple, potentially conflicting objectives that cannot be easily represented by a single scalar reward. Multi-Objective Reinforcement Learning (MORL) addresses this limitation by enabling agents to optimize several objectives simultaneously, explicitly reasoning about trade-offs between them. However, the ``black box" nature of the RL models makes the decision process behind chosen objective trade-offs unclear. Current Explainable Reinforcement Learning (XRL) methods are typically designed for single scalar rewards and do not account for explanations with respect to distinct objectives or user preferences. To address this gap, in this paper we propose \textbf{TREX}, a \textbf{TR}ajectory based \textbf{EX}plainability framework to explain Multi-objective Reinforcement Learning policies, based on trajectory attribution. 

    TREX generates trajectories directly from the learned expert policy, across different user preferences and clusters them into semantically meaningful temporal segments. We quantify the influence of these behavioural segments on the Pareto trade-off by training complementary policies that exclude specific clusters, measuring the resulting relative deviation on the observed rewards and actions compared to the original expert policy. Experiments on multi-objective MuJoCo environments - HalfCheetah, Ant and Swimmer, demonstrate the framework's ability to isolate and quantify the specific behavioural patterns.

\end{abstract}

\section{Introduction}

As the application of Reinforcement Learning for decision-making tasks increases, Explainable AI (XAI) techniques have been extended to Explainable Reinforcement Learning (XRL) with the aim of interpreting agent behavior. Before deploying RL agents in real-world systems, it is crucial to understand the underlying logic behind their decisions to ensure safety and trustworthiness. For example, if an autonomous vehicle engages in sudden braking, it is vital to know if this decision was driven by an external factor, such as an animal crossing, or a sensor failure.

Most complex real-world decision-making domains involve multiple, often conflicting, objectives, requiring the use of Multi-Objective Reinforcement Learning (MORL). In the case of multiple objectives, optimality is defined not by a single policy, but by a set of Pareto-optimal solutions over varying user-preferences. In a typical multi-objective scenario, the RL agent is provided with a user-preference vector that tells the agent how it is expected to priotize the objectives when making decisions. For example, in a task with two objective, a user-preference of (0.5,0.5) denotes both objectives are given the same importance, whereas (0.75, 0.25) tells the agent to favour objective 1 over objective 2 in the given ratio. With MORL policies, explanations must account for learned objective trade-offs and user-specified preferences.  Relative to single objective XRL methods, current literature on Explainable Multi-Objective Reinforcement Learning (XMORL) techniques is scarce \cite{roseroexplainable}. The existing literature primarily focuses on methods such as reward decomposition and policy selection (e.g:\cite{juozapaitis2019explainable}, \cite{tamura2024learning}), resulting in a lack of methods capable of explaining the behaviour of MORL policies.

 In this paper, we introduce \textbf{TREX}, a \textbf{TR}ajectory based \textbf{EX}plainability framework, an explainability framework specifically developed for multi-objective reinforcement learning tasks, and demonstrate its applicability in standard MORL environments from the MO-gymnasium library \footnote{https://mo-gymnasium.farama.org/}.
TREX aims to discover influential behaviours learned by an expert MORL policy using trajectory analysis techniques. It generates trajectories from the expert agent across different user-preferences, with the notion that these trajectories reflect the behavioural patterns learned by the model. For each user-preference we decompose the trajectories into temporal segments and cluster them, essentially grouping semantically similar behaviour patterns in the trajectories. We are then able to measure how these behaviours contribute towards each objective and shape the trade-off at a given user-preference.

For example, in the multi-objective HalfCheetah, a MuJoCo environment from MO-gymnasium where an agent has to optimize for speed vs energy of the robot, certain behaviours drive conflicting outcomes - `long leaps' maximize the robot's speed at the cost of increased energy consumption, while `short strides' conserve energy by sacrificing speed. Consequently, when a user prioritizes energy, speed centric behaviours such as `leaping'  negatively impact the preferred trade-off. Our framework moves beyond just qualitative observations to provide quantitative analyses of these behavioural patterns. The analyses are based on training policies that exclude specific clusters and we measure the relative deviations in the objective returns and objective trade-off, compared to an expert policy. These metrics serve as a proxy for influence, allowing us to pinpoint exactly which behaviours drive objective fulfilment and determining the magnitude of their impact on the Pareto trade-off.

Overall, the contributions of this paper are as follows:

\begin{itemize}
    \item We propose \textbf{TREX}, a post-hoc explainability framework for multi-objective RL, through the behaviours learned by the agent. 
    \item We show how \textbf{TREX} can help quantitatively identify influential behavioural patterns learned by the policy. We then show how these behaviours contribute towards each objective and balance the trade-offs to accommodate a given user-preferences.
\end{itemize}

\section{Related Work}

In this section we provide a brief overview of the relevant background and related work to contextualize our contribution. We first introduce reinforcement learning as a machine learning paradigm (section \ref{Related:RL}) and its multi-objective counterpart (section \ref{Related:MORL}), followed by an overview of existing approaches for explainability in reinforcement learning (section \ref{Related:XRL}), with special focus on the trajectory based approaches (section \ref{Related:T-XRL}). Finally, we introduce the gap in explainability on Multi-Objective Reinforcement Learning approaches (section \ref{Related:XMORL}).

\subsection{Reinforcement Learning}\label{Related:RL} 

Reinforcement Learning (RL) is a machine learning framework in which an agent learns through trial-and-error interactions with its environment, where the Agent iteratively selects an action and receives feedback in the form of rewards or penalties \cite{Sutton1998RL}. By iteratively interacting with the environment, the agent can autonomously acquire policies that achieve optimal or near-optimal behavior without requiring explicit supervision. This learning paradigm achieved contributions in a wide range of domains, including video games \cite{badia2020Outperforming}, traffic signal control \cite{kusic2021spatial}, computer networks \cite{omoniwa2022energy}, robotics \cite{gurtler2023real}, and healthcare \cite{tejedor2020reinforcement}, to mention a few.

\subsection{Multi-Objective Reinforcement Learning}\label{Related:MORL}  

Multi-Objective Reinforcement Learning (MORL) extends standard RL to settings in which agents must simultaneously optimize multiple, potentially conflicting objectives. A common distinction in MORL methods is based on the number of policies learned. In Single-policy approaches the aim is to learn one policy that balances all objectives, most often by reducing the multi-objective problem to a scalar one by the use of methods like reward scalarization or utility functions. Representative examples include weighted-sum methods \cite{abels2019WeigthedSum} and modified Q-learning architectures \cite{li2018ModQN}. In contrast, multi-policy approaches seek to learn a set of policies corresponding to different trade-offs between objectives. These methods aim to explore the Pareto front and allow policies to be selected or adapted based on user preferences or contextual requirements. Examples include policy adaptation methods \cite{yang2019PolicyAdaptation}, that learn and generalize across multiple preference settings, as well as preference-conditioned policies such as Pareto-conditioned networks \cite{reymond2024PCN}. See \cite{hayes2021practical} for a comprehensive survey on MORL.

\subsection{Explainable Reinforcement Learning}\label{Related:XRL}  

A commonly used taxonomy in the Explainable Reinforcement Learning (XRL) literature categorizes explainability techniques according to when explanations are generated into intrinsic and post-hoc methods, and can be further grouped into global or local approaches based on the scope of the explanations \cite{adadi2018peeking,puiutta2020explainable}.  

Intrinsic explainability refers to agents that are constructed to be inherently interpretable by design. These methods restrict the model's complexity to structures that humans can naturally understand (eg: decision trees, programmatic policies). Post-hoc explainability involves generating explanations after an agent has already been trained. In this approach, the agent typically uses a complex, opaque model (eg: Deep Neural Network) to achieve high performance, and a separate method is applied afterwards to interpret its behaviour. Global methods, such as policy-level explanations (eg: \cite{topin2019generation_global_posthoc}), focus on explaining the overall behavioural patterns of the agent. Local techniques focus on explaining specific decisions made by the policy. For example, counterfactual methods explain an agent's decision-making by modelling alternative scenarios, specifically addressing why a particular action was chosen over another \cite{gajcin2024redefining}. For comprehensive surveys on XRL methods, see \cite{milani2024explainable,puiutta2020explainable,vouros2022explainable}

\subsection{Trajectory based Explainable Reinforcement Learning}\label{Related:T-XRL} 

Data attribution is a technique emerged with supervised models, to identify the independent data points that are responsible for the output of the model (eg: \cite{pruthi2020estimating}). In traditional machine learning models, the output is dependent on the individual data points, but a RL agent's decision is typically a results of the sequence of past experiences. Therefore, in XRL, typically trajectories are used to interpret model behaviours and provide explanations. 

HIGHLIGHTS \cite{amir2018highlights} propose a technique to summarise policy behaviour to humans, by identifying important trajectories and states during an agent's execution. Summaries are provided as trajectories, identified by discovering \textit{`importance states'} and then extracting the sequence of states and actions neighbouring the significant state in order to create the trajectory. Studies on trajectory-aware method such as \cite{deshmukh2023explaining} finds influential trajectories in the training dataset, that are responsible for decisions made by the policy. This is done by iteratively training smaller policies with specific trajectories removed from the training data and attributing the changed behaviours to the missing trajectories.

\subsection{Explainable Multi-Objective Reinforcement Learning} \label{Related:XMORL} 

Explainability in MORL (XMORL) is yet to be explored, and has many challenges \cite{roseroexplainable}, existing works on explainable MORL is primarily focused on policy selection. Unlike single objective RL which yields a single optimal policy, a MORL algorithm has to balance multiple objectives, which can result in multiple possible optimal policies. Therefore, MORL methods usually provide a set of solutions (i.e Pareto set), in which each solution represents a distinct trade-off among the objectives. 
While these MORL methods successfully navigate complex trade-offs, they impose significant challenges in understanding the underlying decision making logic. Consequently, to explain how the policy decisions and resulting behaviours contribute to different objectives with varying user preferences, explainability techniques are required.

Typically when selecting a policy out of the optimal set, the user or decision maker looks at the objective values/returns to select a policy based on their preferences for specific objectives. Tamura et. al. \cite{tamura2024learning} propose a penalty metric "Mismatch" - determine the disparity between the user's preferences and the solution set. By minimizing the "Mismatch", it helps the users to select policies from the Pareto set that aligns with the user's preferences. Osika et. al. \cite{osika2024navigating} propose a technique to help decision makers or the user select a policy from the solution set by summarizing MORL solution set using both objective returns and behavioural space (i.e: using the HIGHLIGHT \cite{amir2018highlights} to identify important states). 

Existing XMORL approaches are scarce, and focus on providing support for policy selection by summarizing objective returns or highlighting salient states, but they offer limited insight into the underlying behaviours that drive objective trade-offs within a policy. Current methods do not explain how distinct behavioural patterns contribute to different objectives, or how these behaviours change across user preferences. As a result, they fall short of explaining the different trade-off logics learned by multi-objective policies.  Our approach addresses this by attributing objective trade-offs to semantically meaningful behavioural patterns extracted directly from the learned policy.

\section{Methodology}

In this section we introduce \textbf{TREX}, a \textbf{TR}ajectory based \textbf{EX}plainability framework to explain Multi-objective Reinforcement Learning policies using post-hoc techniques. 

TREX is designed to explain what behaviours the expert MORL agent has learned and how the behaviours change across different trade-off preferences, in order to make decisions that suit a given user-preference. The primary analysis approach of TREX is \textbf{Preference Level Analysis (PLA)}. In this analysis we quantitatively identify learned behaviours of the expert policy at a given user-preference, discovering distinct behaviour sequences, measuring their influence on each objective and overall objective trade-offs.

\subsection{Preference-Level analysis (PLA)}
\label{sec:preference_level}

A multi-objective reinforcement learning agent typically learns a set of Pareto optimal solutions over a range of user-preferences. When deployed, the agent makes decisions to fulfil the expected objective outcomes of the user - specified by the user-preference utility. Therefore the agent's decisions and the resulting behaviours can vary from one preference to another. 

In this section, we focus on analysing the behaviours at individual single user preferences. The intuition is that, in an episodic trace, there can be different behavioural segments observed from the actions taken by the agent. These reflect how the agent learned to make decisions in order to optimize each objective and to accommodate a specific user preference, i.e. a specific trade-off ratio between the objective. The significance of these behaviours on the individual objectives and on the overall objective trade-off can vary. To measure these quantitatively, we conduct a trajectory analysis by adapting the notion of trajectory attribution from \cite{deshmukh2023explaining}. 

In PLA, we use attribution techniques to determine the contribution of certain trajectories towards the agent's learned behaviour. The intuition behind this approach is that if we train two different policies - one trained by removing a set of trajectories from the training data (i.e: complementary policy) and another with the complete training data (i.e: original policy) - the shift in the distribution relative to the original policy is an indication of the importance of the missing trajectories. This is the underlying notion used in PLA.

\begin{figure}[!h]
\includegraphics[width=\textwidth]{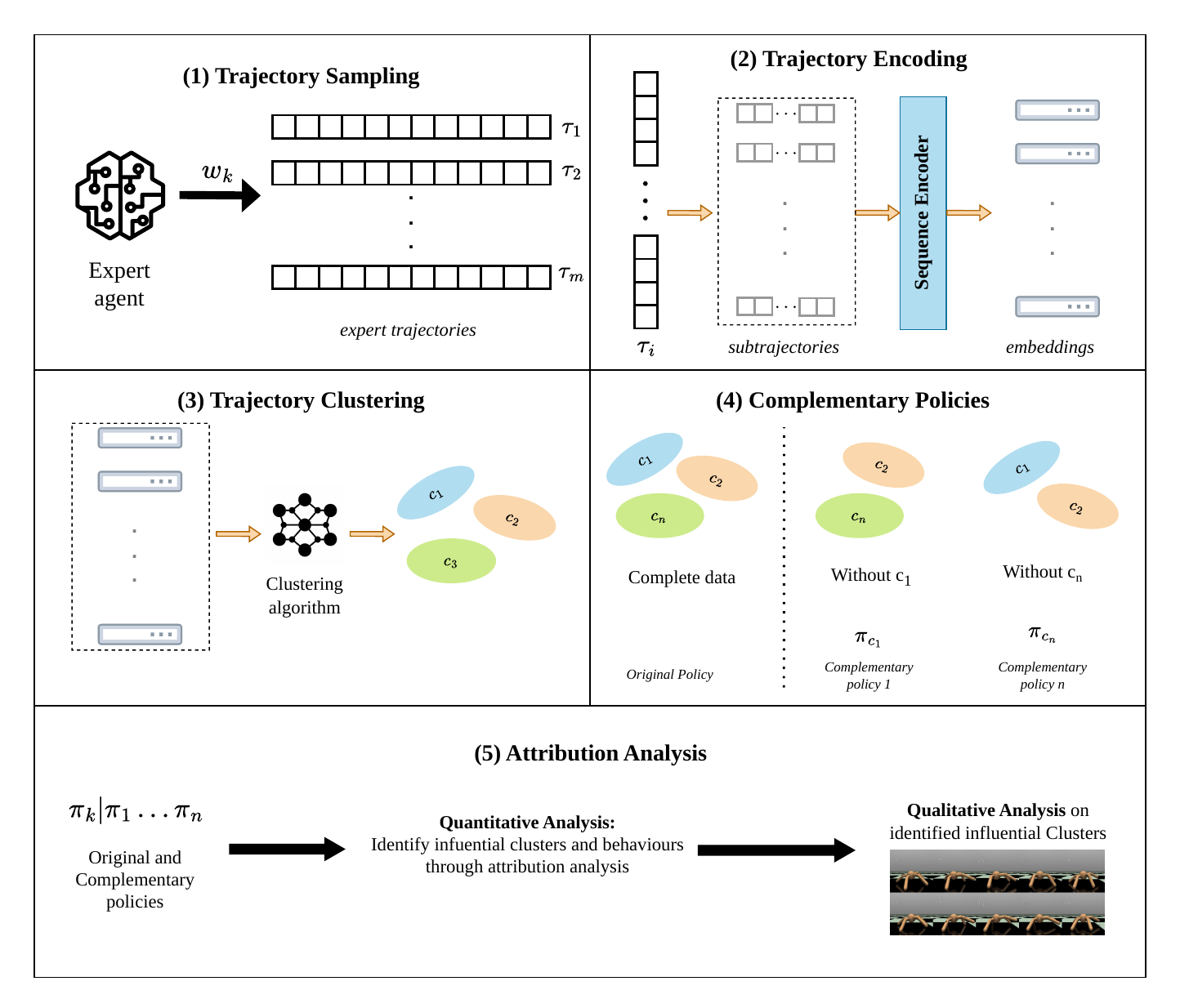}
\caption{Overview of the Preference-Level Analysis (PLA) pipeline: (1) sampling trajectories from an expert agent, (2) encoding sub-trajectories into latent space embeddings, (3) clustering trajectories into semantically distinct behaviours, (4) training original policy alongside complementary policies that iteratively exclude specific cluster trajectories, and (5) conducting attribution analysis to quantify the behavioural influence of each cluster.}
\label{fig:pref_level}
\end{figure}

The preference level analysis is conducted in 5 steps (summarised in Fig.\ref{fig:pref_level}.)

\begin{enumerate}
    \item \emph{Trajectory Sampling:} We take a set of user-preferences ${\mathcal{W} = \{\omega_1, \omega_2, \cdots \omega_{k} \}}$ and for each preference $\omega_{k}$ we generate an offline dataset $D_k$ using trajectories generated using the expert policy $\pi_E$. At each step $t$, the agent takes a random action with probability $\epsilon$ and follows the learned policy with probability $(1-\epsilon )$. Each dataset consist of $m$ episodes with a maximum length of $T$. 

    \item \emph{Trajectory encoding:} The trajectories are split into marginally overlapping sub-trajectories of each length $l$, which is a timely sequence of observations, actions and rewards. Each sub-trajectory is then encoded into a latent representation (i.e trajectory embedding) using a sequence encoder.

    \item \emph{Trajectory Clustering:} The embeddings of the sub-trajectories are then clustered to a set of clusters $\mathbf{C}_k$. We use K-means algorithm for clustering and the number of clusters $n_k$ are determined using silhouette analysis. 

    \item \emph{Complementary policies: } For each preference $\omega_{k}$, an original policy $\pi_k$ is trained using the full dataset $\mathbf{D}_k$ which consist of trajectories in all the clusters. 
    We utilize $\pi_k$ as a surrogate for the expert policy $\pi_E$ that mimics the expert agent. As $\pi_k$ is trained solely on the offline dataset $D_k$ generated by the $\pi_E$, its learned distribution acts as a proxy for the expert agent’s decision-making logic.
    Then a set of complementary policies are trained, where each complimentary policy $\pi_c$ is trained by removing trajectories belonging to cluster $c$. Note than only one cluster is removed at a time. 

    \item \emph{Attribution Analysis: } To quantitatively identify influential behavioural clusters within $\omega_k$, we conduct an attribution analysis explained in the following section \ref{sec:attribution_analysis}.
\end{enumerate}

\subsection{Attribution Analysis}
\label{sec:attribution_analysis}

To quantitively determine the influence of the trajectories on the agent's decision, we analyse the returns obtained by each of the trained policies. 

\emph{Reward Attribution Score: } For each objective $i \in I$, we measure relative change in the objective returns $\Delta R^i$ from each complementary $\pi_c$ policy (in Eq.\ref{eq:delta_ri}), relative to the returns of the original policy $\pi_k$. $R^i_k$ indicates the cumulative return for objective $i$ using the original policy $\pi_k$ and $R^i_c$ is the cumulative return for complementary policy $\pi_c$. 

\begin{equation}
    \Delta R^i=(R^i_k - R^i_{c})/R^i_k \label{eq:delta_ri}
\end{equation} 

We then calculate the total returns deviation $\Delta R$ in Eq.\ref{eq:delta_rc}, which is the magnitude of the combined deviation of all the objective returns - calculated as a L2 norm distance. $\Delta R^i$ is the relative change in objective returns for objective $i$.

\begin{equation}
    \Delta R \left( c \right) = \sqrt{ \sum_{i \in I} \left( \Delta R^i \right)^2 } \label{eq:delta_rc}
\end{equation}

Take an instance where the complementary policy's observed returns diminish for all objectives simultaneously. This drop in performance may be more indicative of poor training data rather than the inherent significance of that specific cluster. A more robust metric for determining the influence of a cluster on objective trade-offs involves the observation of contrary deviations. Meaning, if the exclusion of a set of trajectories yields improved performance in one objective alongside a decrease in others, it may highlight the cluster's pivotal role in the agent's learned trade-off logic. Such inverse reward shifts provide quantitative evidence of how distinct behaviours are utilized to navigate the competing demands of the Pareto front.

Therefore we calculate the Reward Attribution Score for each cluster, which is a measure the cluster's influence on objective trade-offs. Eq.\ref{eq:reward_attribution} calculates the $RAS(c)$ for 2-objectives.

\begin{equation}
    RAS \left( c \right) = \left| w_1 \Delta R^1 - w_2 \Delta R^1 \right| \label{eq:reward_attribution}
\end{equation} 

where $w_1, w_2$ is the preference on each objective (i.e $\omega_k = (w_1, w_2)$. For more than 2 objectives, Eq.\ref{eq:reward_attribution} can be extended to calculate $RSA$ in a combinatorial manner. A higher $RAS$ score depicts a higher shift in the objective trade-off.

\section{Experimental Setup}

\begin{figure}[!h]
     \centering
     \begin{subfigure}[b]{0.25\textwidth}
         \centering
         \includegraphics[width=\textwidth]{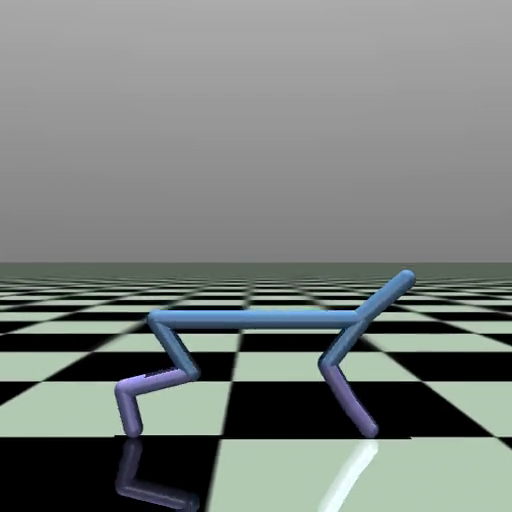}
         \caption{MO-HalfCheetah}
     \end{subfigure} 
     \hspace{2mm}  
     \begin{subfigure}[b]{0.25\textwidth}
         \centering
         \includegraphics[width=\textwidth]{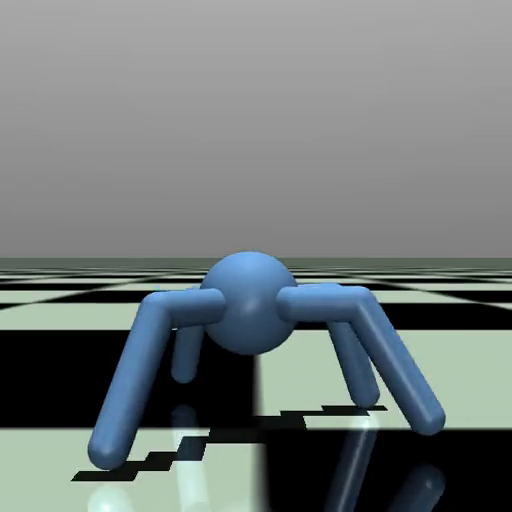}
         \caption{MO-Ant}
     \end{subfigure}%
     \hspace{2mm}
     \begin{subfigure}[b]{0.25\textwidth}
         \centering
         \includegraphics[width=\textwidth]{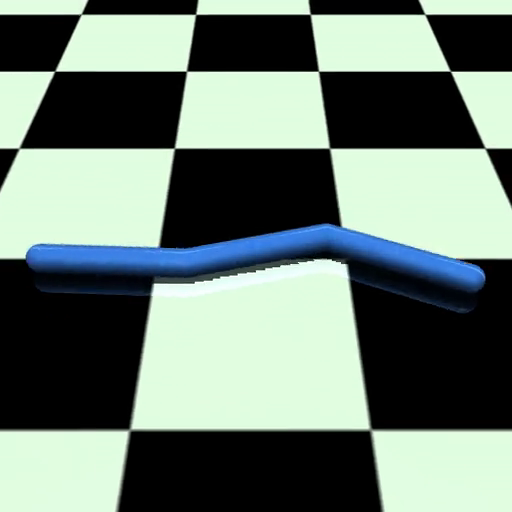}
         \caption{MO-Swimmer}
     \end{subfigure}
     \caption{MuJoCo multi-objective environments}
     \label{fig:environments}
\end{figure}

\textbf{Environments:} We conduct our experiments on three multi-objective MuJoCo environments from MO-Gymnasium\footnote{https://mo-gymnasium.farama.org/environments/MuJoCo/}, namely (1) MO-HalfhCheetah (2) MO-Ant and (3) MO-Swimmer, shown in Fig.\ref{fig:environments}.

\begin{itemize}
    \item MO-HalfCheetah: In the Halfcheetah environment, an agent controls a 2-limb robot with two conflicting objectives - maximising the robot speed and minimizing the energy consumption. 
    \item MO-Ant: In the MO-Ant environment, the agent controls a 4-legged robot with 8 actuated joints and has two objectives - distance in x-axis and y-axis respectively. 
    \item MO-Swimmer: The environment involves an agent controlling a multi-linked robot designed to swim in a 2D fluid by coordinating its joints the agent controls. The two objectives for this task are speed and energy saving. 
\end{itemize}

\noindent
\textbf{Expert Policy: } We use a preference conditioned Multi-Objective Decision Transformer \textit{MODT(P)} from the PEDA framework \cite{zhu2023scaling} as our expert policy. The MODT(P) is trained using the offline D4MORL dataset provided by the authors of PEDA. We use the same parameters provided by the authors.

\noindent
\textbf{Complementary Policies: } In preference-level analyses (section \ref{sec:preference_level}), we use a smaller scale Multi-Objective Decision Transformer to train the original $\pi_k$ and complimentary policies $\pi_c$. The MODT can be replaced with any suitable multi-objective RL agent. During training of each policy, we select the model checkpoints that give the best weighted/scalarised return, with respect to the preference utility $\omega_k$, and are then used in our evaluations. 

\noindent
\textbf{Preference Level Analysis: } For this analysis, we use the set of user-preferences $W$ = \{[0.25, 0.75], [0.5, 0.5], [0.75, 0.25]\}. 
To generate the offline dataset for each preference, we run $m=25$ episodes of simulations with $\epsilon = 0.05$. Each episode is run for a maximum of $T=500$ time-steps or until terminated. The trajectories are then split into sub-trajectories of length $l=20$ with an overlap margin of $\alpha=5$. To encode the sub-trajectories, we utilize the attention-based transformer of the trained expert MODT(P) as the sequence encoder. The sequence of \textit{state-action-return} triplets are input to the model and the output tokens from the final transformer block is taken and averaged to generate the trajectory embedding. Our framework is flexible as to use any suitable sequence encoder for this. 

\noindent
\textbf{Code: } The implementation of our TREX framework is provided in the github repository \url{https://github.com/dilina-r/trex_xmorl}.

\section{Results}

In this section we evaluate the proposed TREX framework through quantitative and qualitative analysis across multiple multi-objective environments. We examine the influence of behavioural clusters under different user-preference settings and qualitatively explain influential trajectories identified.

\subsection{Preference Level Analysis - Quantitative Results}

Tables \ref{tbl:attribution_halfcheetah}--\ref{tbl:attribution_swimmer} show the preference level analysis results for MO-HalfCheetah, MO-Ant and MO-Swimmer environments respectively. The tables show the cumulative returns $R^i$ obtained by the expert agent, original policy and complementary policies, across three user-preference settings. The tables also show the returns deviation $\Delta R^i$ for each objective, the total return deviation $\Delta R(c)$ and the Reward Attribution score $RAS$ for each complementary policy $\pi_c$. The most influential trajectory cluster (i.e highest $RAS$ score) for each preference utility is highlighted in bold.

Firstly, we observe the original policy achieves returns closer to the expert, but in several instances it is seen to be notably different. This could potentially be due to the disparity between the dataset and training process. We take that the original policy, which is trained on the expert trajectories, is able to sufficiently learn the behaviours of the expert agent.

\begin{table}[!h]
\centering
\caption{Preference Level analysis results for MO-HalfCheetah environment - Best overall return deviation $\Delta R$ and Reward attribution score for each preference is highlighted in bold. }
\label{tbl:attribution_halfcheetah}
\begin{tabular}{|c|c|c|c|c|c|c|c|}
\hline
\textit{preference} & \textit{policy} & $R^1$ & $R^2$ & $\Delta R^1$ & $\Delta R^2$ & $\Delta R \left( c \right)$ & \textit{RAS(c)} \\ \hline
\multirow{5}{*}{(0.25,0.75)}  & expert          & 1033.403    & 2417.319    & -              & -              & -              &                 \\ \cline{2-4}
                              & original        & 917.385     & 2419.520    & -              & -              & -              &                 \\
                              & c0              & 769.243     & 2440.969    & 0.161          & -0.009         & \textbf{0.162} & \textbf{0.170}  \\
                              & c1              & 780.596     & 2435.989    & 0.149          & -0.007         & 0.149          & 0.156           \\
                              & c2              & 917.432     & 2414.269    & 0.000          & 0.002          & 0.002          & 0.002           \\ \hline
\multirow{5}{*}{(0.5,0.5)}    & expert          & 1975.156    & 2237.225    & -              & -              & -              &                 \\ \cline{2-4}
                              & original        & 1781.191    & 2128.915    & -              & -              & -              &                 \\
                              & c0              & 2386.647    & 1813.144    & -0.340         & 0.148          & \textbf{0.371} & \textbf{0.488}  \\
                              & c1              & 2206.210    & 2004.030    & -0.239         & 0.059          & 0.246          & 0.297           \\
                              & c2              & 1272.957    & 2359.837    & 0.285          & -0.108         & 0.305          & 0.394           \\ \hline
\multirow{8}{*}{(0.75, 0.25)} & expert          & 2442.122    & 1740.961    & -              & -              & -              &                 \\ \cline{2-4}
                              & original        & 2423.394    & 1717.668    & -              & -              & -              &                 \\
                              & c0              & 2417.976    & 1772.944    & 0.002          & -0.032         & 0.032          & 0.034           \\
                              & c1              & 2418.387    & 1665.930    & 0.002          & 0.030          & 0.030          & 0.028           \\
                              & c2              & 2413.507    & 1749.870    & 0.004          & -0.019         & 0.019          & 0.023           \\
                              & c3              & 2410.540    & 1798.680    & 0.005          & -0.047         & \textbf{0.047} & \textbf{0.052}  \\
                              & c4              & 2430.640    & 1727.198    & -0.003         & -0.006         & 0.006          & 0.003           \\
                              & c5              & 2381.938    & 1736.558    & 0.017          & -0.011         & 0.020          & 0.028           \\ \hline
\end{tabular}
\end{table}

The preference-level analysis reveal that specific behavioural clusters significantly influence the objective outcomes and the agent's ability to navigate objective trade-offs. 

In the MO-HalfCheetah environment, removing certain clusters led to distinct shifts in the Pareto front. For example, at the preference $(0.5, 0.5)$, removing Cluster 0 resulted in a negative deviation for speed ($\Delta R^1$) but a positive deviation for energy conservation ($\Delta R^2$), effectively shifting the policy's priority. Conversely, removing Cluster 2 in the same setting shifted the trade-off to favour Objective 2. This behaviour is reflected in the high attribution scores $RAS(c)$ for the relevant clusters.

\begin{table}[!h]
\centering
\caption{Preference Level analysis results for MO-Ant environment. Best overall return deviation $\Delta R$ and Reward attribution score for each preference is highlighted in bold. }
\label{tbl:attribution_ant}
\begin{tabular}{|c|c|c|c|c|c|c|c|}
\hline
\textit{preference} & \textit{policy} & $R^1$ & $R^2$ & $\Delta R^1$ & $\Delta R^2$ & $\Delta R \left( c \right)$ & \textit{RAS(c)} \\ \hline
\multirow{5}{*}{(0.25,0.75)} & expert   & 1007.504 & 2427.701 & -      & -      & -              & -              \\ \cline{2-4}
                             & original & 912.711  & 2385.175 & -      & -      & -              & -              \\
                             & c0       & 777.709  & 938.847  & 0.148  & 0.606  & \textbf{0.624} & \textbf{0.418} \\
                             & c1       & 837.784  & 2307.194 & 0.082  & 0.033  & 0.088          & 0.004          \\
                             & c2       & 915.246  & 2406.504 & -0.003 & -0.009 & 0.009          & 0.006          \\ \hline
\multirow{5}{*}{(0.5,0.5)}   & expert   & 2066.393 & 1940.264 & -      & -      & -              & -              \\ \cline{2-4}
                             & original & 1713.007 & 1904.959 & -      & -      & -              & -              \\
                             & c0       & 1474.368 & 1653.254 & 0.139  & 0.132  & \textbf{0.192} & 0.004          \\
                             & c1       & 1826.586 & 1970.307 & -0.066 & -0.034 & 0.075          & 0.009          \\
                             & c2       & 1871.536 & 1829.054 & -0.093 & 0.040  & 0.101          & \textbf{0.053} \\ \hline
(0.75, 0.25)                 & expert   & 2422.992 & 1249.866 & -      & -      & -              & -              \\ \cline{1-4}
                             & original & 2286.746 & 1347.436 & -      & -      & -              & -              \\
                             & c0       & 2025.750 & 1323.575 & 0.114  & 0.018  & 0.115          & \textbf{0.081} \\
                             & c1       & 2209.261 & 1252.776 & 0.034  & 0.070  & 0.078          & 0.008          \\
                             & c2       & 1999.653 & 1155.578 & 0.126  & 0.142  & \textbf{0.190} & 0.059          \\ \hline
\end{tabular}
\end{table}

In the MO-Swimmer environment, Objective 1 (i.e: distance) was highly sensitive to cluster removal, while Objective 2 (i.e control cost) showed no deviation across all preferences. Furthermore the deviations were predominantly positive, meaning returns decreased upon trajectory removal, which indicates that these segments were vital for maintaining performance. Notably, across both MO-HalfCheetah and MO-Swimmer, in the preference $(0.75, 0.25)$ removal of any cluster failed to produce significant reward deviations.

\begin{table}[!h]
\centering
\caption{Preference Level analysis results for MO-Swimmer environment. Best overall return deviation $\Delta R$ and Reward attribution score for each preference is highlighted in bold. }
\label{tbl:attribution_swimmer}
\begin{tabular}{|c|c|c|c|c|c|c|c|}
\hline
\textit{preference} & \textit{policy} & $R^1$ & $R^2$ & $\Delta R^1$ & $\Delta R^2$ & $\Delta R \left( c \right)$ & \textit{RAS(c)} \\ \hline
\multirow{8}{*}{(0.25,0.75)} & expert          & 51.324      & 149.375     & -              & -              & -               & -              \\ \cline{2-4}
                             & original        & 69.763      & 147.884     & -              & -              & -               & -              \\
                             & c0              & 73.300      & 147.842     & -0.051         & 0.000          & \textbf{0.051}  & 0.013 \\
                             & c1              & 85.556      & 146.778     & -0.226         & 0.007          & 0.227           & 0.062          \\
                             & c2              & 59.052      & 147.240     & 0.154          & 0.004          & 0.154           & 0.035          \\
                             & c3              & 50.273      & 148.138     & 0.279          & -0.002         & \textbf{0.279}  & \textbf{0.071} \\
                             & c4              & 84.131      & 146.982     & -0.206         & 0.006          & 0.206           & 0.056          \\
                             & c5              & 51.083      & 148.268     & 0.268          & -0.003         & 0.268           & 0.069          \\ \hline
\multirow{5}{*}{(0.5,0.5)}   & expert          & 129.946     & 140.608     & -              & -              & -               & -              \\ \cline{2-4}
                             & original        & 100.249     & 142.083     & -              & -              & -               & -              \\
                             & c0              & 31.933      & 148.406     & 0.681          & -0.045         & \textbf{0.683}  & \textbf{0.363} \\
                             & c1              & 73.927      & 143.104     & 0.263          & -0.007         & 0.263           & 0.135          \\
                             & c2              & 125.910     & 141.672     & -0.256         & 0.003          & 0.256           & 0.129          \\ \hline
\multirow{5}{*}{0.75, 0.25}  & expert          & 233.001     & 73.352      & -              & -              & -               & -              \\ \cline{2-4}
                             & original        & 222.898     & 80.444      & -              & -              & -               & -              \\
                             & c0              & 216.077     & 82.804      & 0.031          & -0.029         & \textbf{0.042}  & \textbf{0.030} \\
                             & c1              & 219.761     & 83.444      & 0.014          & -0.037         & 0.040           & 0.020          \\
                             & c2              & 217.289     & 82.123      & 0.025          & -0.021         & 0.033           & 0.024          \\ \hline
\end{tabular}
\end{table}

In the MO-Ant environment, it is noteworthy that a higher $\Delta R(c)$ does not always result in a higher attribution score $RAS(c)$, as seen with the preferences (0.5,0.5) and (0.75,0.25). The higher $\Delta R(c)$ is due to positive deviations in the returns, meaning the returns are reduced for both objectives simultaneously. While this can suggest the significance of the trajectories in the removed clusters on the performance of both objectives, it could also imply a lack of diverse training data.

\subsection{Preference Level Analysis - Qualitative Results}

The quantitative results from preference level analysis discover influential clusters and provides insights into how they affect the objectives. To validate the findings in the quantitative experiments, we conduct a qualitative analysis by visualising the agent behaviour in each cluster. For this analysis, we visualise the best trajectories of each cluster, which are the trajectories closest to the centroid of a cluster. We observe the behaviours in these visual segments and correlate them to the significant clusters in the quantitative results. 

\begin{figure}[!h]
     \centering
     \begin{subfigure}[b]{0.75\textwidth}
         \centering
         \includegraphics[width=\textwidth]{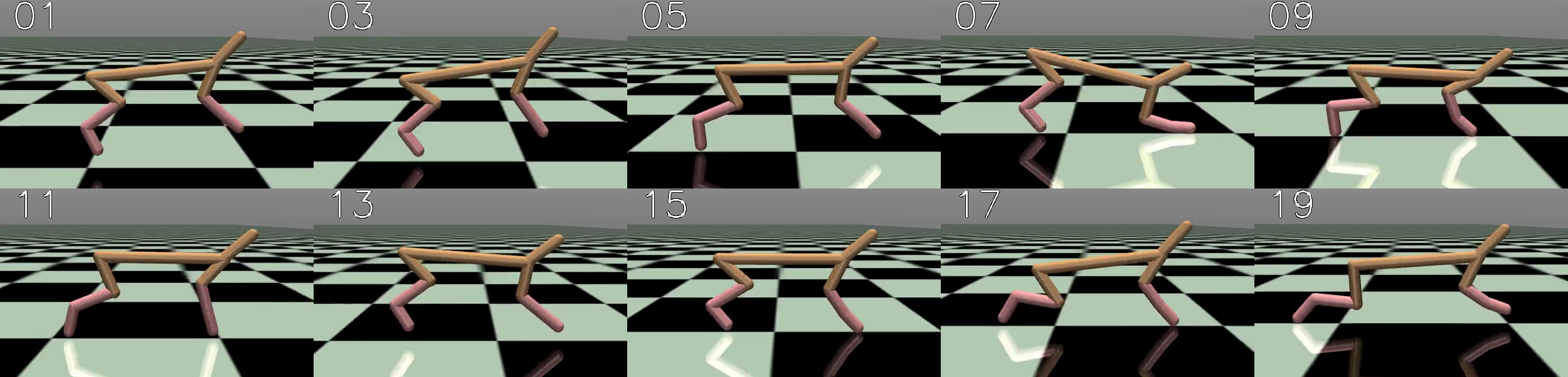}
         \caption{Cluster 0}
         \label{fig:top}
     \end{subfigure}
     
     \begin{subfigure}[b]{0.75\textwidth}
         \centering
         \includegraphics[width=\textwidth]{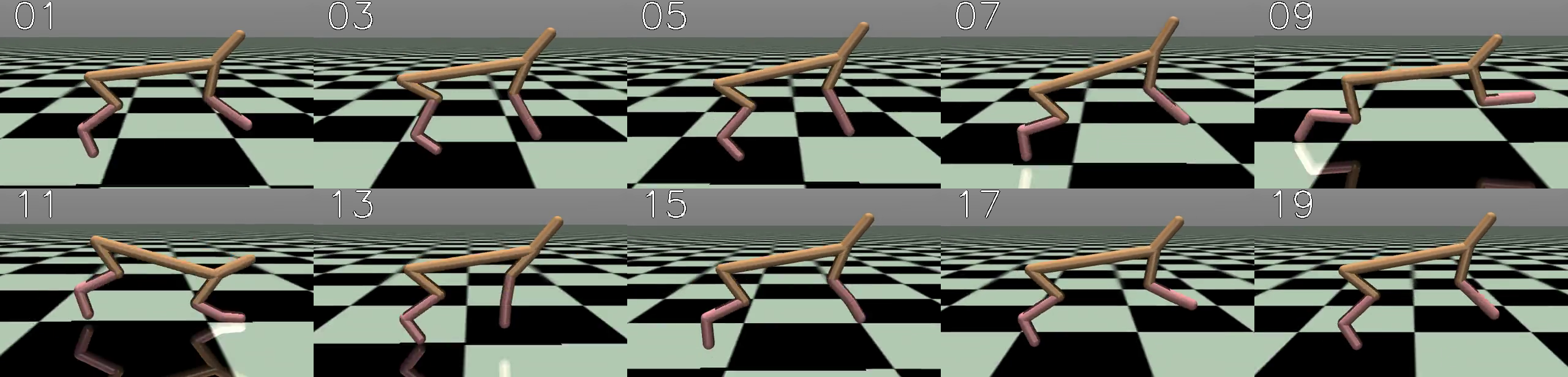}
         \caption{Cluster 1}
         \label{fig:middle}
     \end{subfigure}

     \begin{subfigure}[b]{0.75\textwidth}
         \centering
         \includegraphics[width=\textwidth]{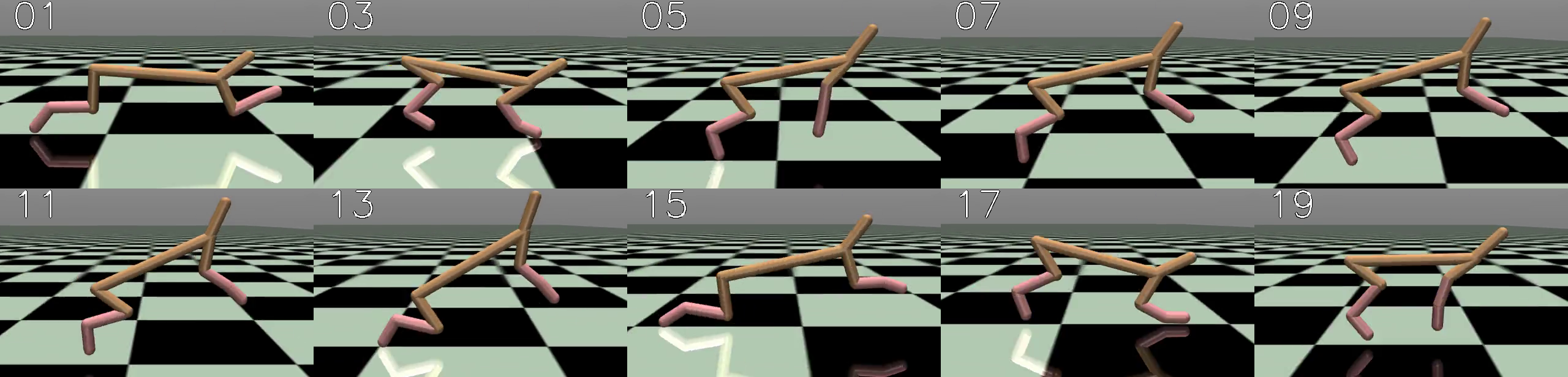}
         \caption{Cluster 2}
         \label{fig:bottom}
     \end{subfigure}
     
     \caption{MO-HalfCheetah trajectories for the clusters with (0.5, 0.5) user-preference}
     \label{fig:halfcheetah_visualise1}
\end{figure}

We first look at MO-HalfCheetah at (0.5, 0.5) preference, where the highest deviations are seen. Fig.\ref{fig:halfcheetah_visualise1} illustrates the visual sequence of the trajectory in each cluster. Quantitative results in Table \ref{tbl:attribution_halfcheetah} indicate that the exclusion of Cluster 0 yields improved \textit{speed} alongside decreased \textit{energy saving}. According to the attribution analogy, it suggests that Cluster 0 should exhibit behaviours that oppose the \textit{speed} objective. Visual inspection support this hypothesis, showing that in Cluster 0, the robot takes smaller, slower hops compared to the more expansive, high velocity strides of Clusters 1 and 2. Notably, the omission of the highest performing cluster (Cluster 2) leads to an inverse effect, reducing returns on the \textit{speed} objective.

\begin{figure}[!h]
     \centering
     \begin{subfigure}[b]{0.75\textwidth}
         \centering
         \includegraphics[width=\textwidth]{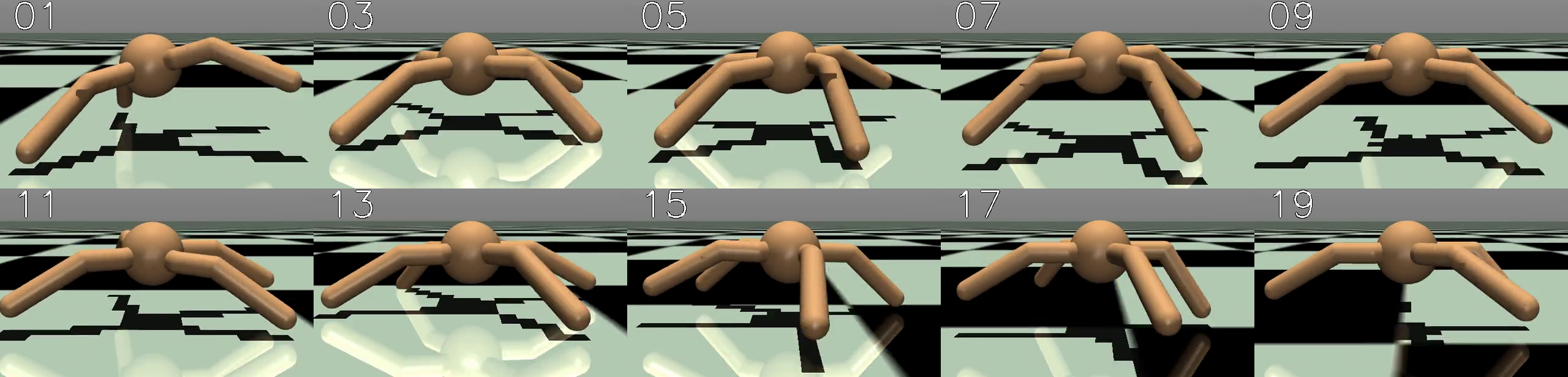}
         \caption{Cluster 0}
         \label{fig:top}
     \end{subfigure}
     
     \begin{subfigure}[b]{0.75\textwidth}
         \centering
         \includegraphics[width=\textwidth]{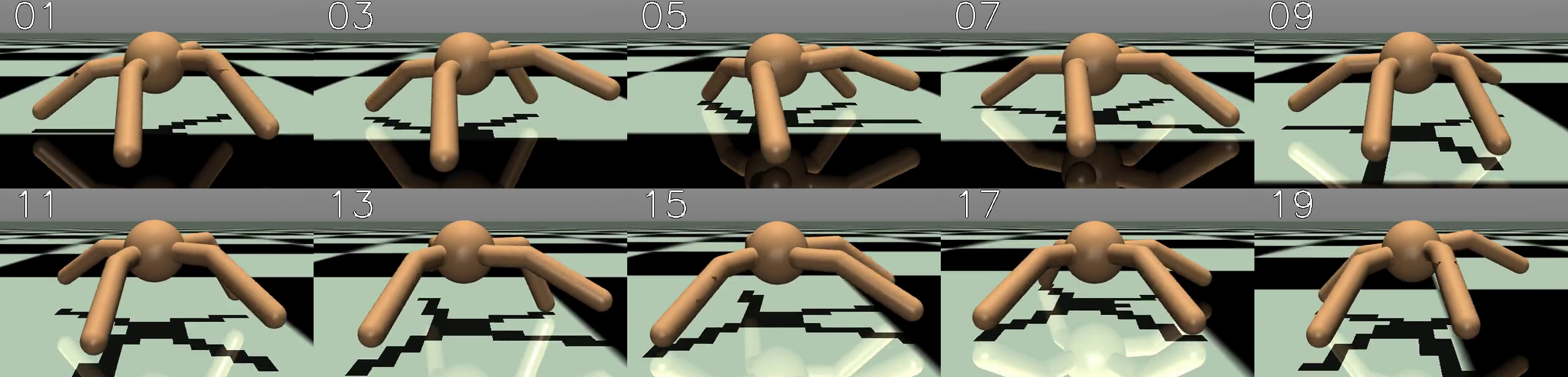}
         \caption{Cluster 1}
         \label{fig:middle}
     \end{subfigure}

     \begin{subfigure}[b]{0.75\textwidth}
         \centering
         \includegraphics[width=\textwidth]{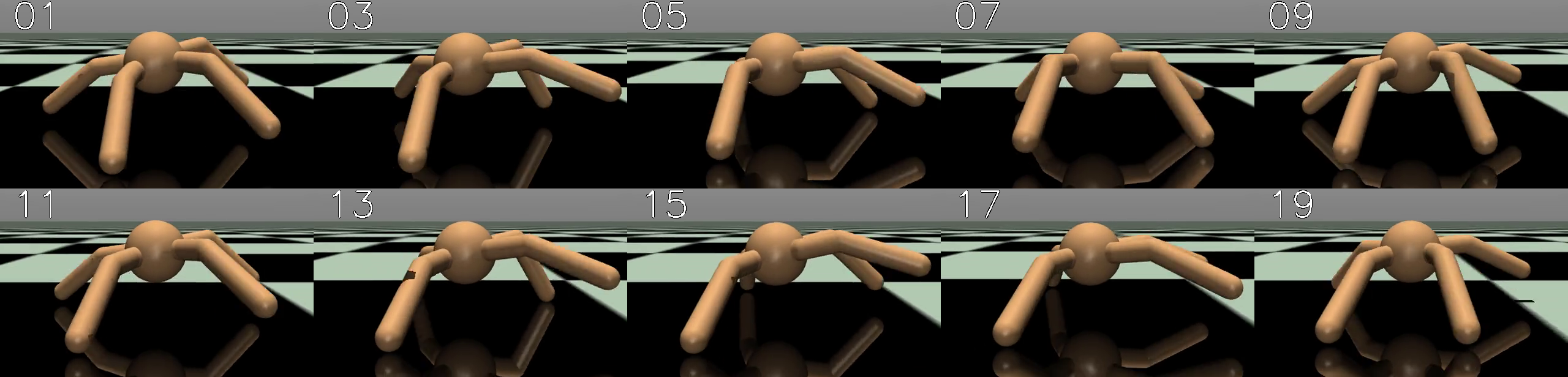}
         \caption{Cluster 2}
         \label{fig:bottom}
     \end{subfigure}
     
     \caption{MO-Ant trajectories for the clusters with (0.5, 0.5) user-preference}
     \label{fig:ant_050_050_visualise}
\end{figure}

In Table \ref{tbl:attribution_ant} with MO-Ant, it is seen for the preference (0.5, 0.5) removing Cluster 0 reduces both objective returns, suggesting that either the trajectories in Cluster 0 contributes positively towards both objectives simultaneously, or it is simply due to poor training data. Upon observing the motions of each cluster as seen in Figure \ref{fig:ant_050_050_visualise}, it was seen in Cluster 0 the robot moves `diagonally' at a higher velocity compared to Cluster 1 and Cluster 2. Therefore removing Cluster 0 results in poor performance in both objectives, which is the distance in x-axis and y-axis. 

\begin{figure}[!h]
\centering
\includegraphics[width=0.75\textwidth]{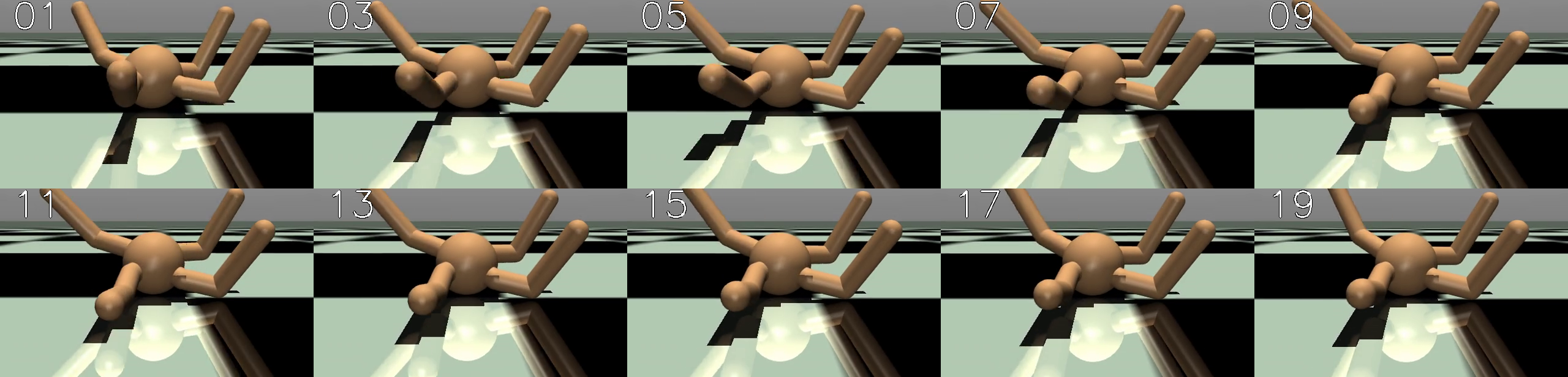}
\caption{MO-Ant Cluster 2 trajectory with (0.25, 0.75) user-preference}
\label{fig:ant_qual_c2}
\end{figure}

The preference (0.25, 0.75) shows the highest returns deviations and $RAS$ score. It was seen that in Cluster 0, the robot moves at a higher velocity leaning in the north direction (y-axis), relative to the remaining clusters. Note that removal of Cluster 2 did not result any significant deviations. As seen in Fig. \ref{fig:ant_qual_c2} in Cluster 2 trajectories the robot is immobile, yielding no returns.

\section{Discussion}

The experimental results demonstrate that our TREX framework successfully identifies and quantifies the behavioural strategies learned by MORL agents across different user preferences. By correlating quantitative attribution scores with qualitative visual evidence, we can validate the findings of the preference level analysis and provide significant insights into how agents navigate complex trade-offs.

The discovery of \textit{contrary deviations} through Reward Attribution Score (RAS) analysis provides the most robust evidence of an agent's internal trade-off logic. In the MO-HalfCheetah (0.5, 0.5) setting, the exclusion of Cluster 0 actually improved energy-saving while decreasing speed, confirming that these specific trajectory segments were primarily responsible for driving the speed objective at the cost of efficiency. Our quantitative results also highlighted scenarios where removing a cluster causes all objectives to diminish simultaneously, as seen in some MO-Ant preference settings. Initially this brought the question whether the observed deviations indicate a lack of diverse training data rather than a significant behavioural segment. Through our qualitative analysis, we verified that the cluster indeed improves the outcomes of both objectives, hence removing it causes a drop in overall performance. 

The findings of the experiments also provide insights into the nature of the environments and how the expert agent has learned to make decisions to navigate the objective trade-off in each environment. In tasks with conflicting objectives, such as in HalfCheetah and Swimmer, the quantitative results show that the deviations in returns only occur in opposing directions - meaning both objectives do not show significant increase or decrease simultaneously. On the other hand, in the Ant environments it was observed that the objectives are not always strictly conflicting. It was depicted with observations for (0.5, 0.5) user-preference that certain cluster behaviours can contribute towards both objective returns. This suggests that the agent learned to exploit the geometry of the environment to optimize both simultaneously. 

Furthermore, we see that objective 2 in MO-Swimmer appears to be less sensitive to specific behavioural variations in this environment. The results showed high sensitivity for Objective 1 (i.e: distance) but almost no deviation for Objective 2 (i.e: energy or control cost). This can suggest that the cost of movement of the might be more uniform across the behaviours within a preference setting.

\subsection{Limitations and Future work}

Our framework offers a foundation for building trust in multi-objective systems by revealing specific behavioural components that justify its trade-offs. However we identify few limitations of our evaluations, and avenues for further refinement of the framework. 

\textbf{Proxy models: } One of the observed limitation in our framework is that the preference-level analysis relies on the original policy to closely mimic the expert agent at a given preference. The disparity observed between the expert policy returns and the original policy returns in some instances (seen in Tables \ref{tbl:attribution_halfcheetah}--\ref{tbl:attribution_swimmer}) suggests that the mimicking process (training $\pi_k$ on expert trajectories) may introduce minor behavioural shifts. Future work could focus on refining this distillation process to ensure the explanations are as faithful to the expert as possible. 

\textbf{Computational cost:} A practical limitation of the TREX framework is the computational cost associated with the training of the original and complementary policies in the attribution analysis. For every identified behavioural cluster, a new complementary policy $\pi_c$ must be trained, meaning the computational overhead scales linearly with the number of clusters. To alleviate the computational burden, we train these policies using less complex RL models. Even with this mitigation, the post-hoc analysis phase remains computationally intensive. However, the existing architecture of the framework is necessary to provide highly granular, quantitative explanations of the agent behaviour.

\textbf{Encoding and clustering:} Currently our framework is sensitive to the choice of trajectory encoding and clustering. Fig.\ref{fig:halfcheetah_clusters}--\ref{fig:swimmer_clusters} in appendix \ref{appendix_a} illustrate the clusters of the MO-Halfcheetah, MO-Ant and MO-Swimmer obtained with K-means clustering, under three preference settings. The figures depict that the sub-trajectories show good separation among the clusters. However some of the clusters have considerably large boundaries and there could be multiple less frequent behaviours within the same cluster. For example, under the preference (0.25, 0.75) in MO-Ant (see Fig.\ref{fig:ant_pref1}), Cluster 2 has two or more visible blobs of data-points within the same cluster. Note that the number of clusters are determined using the silhouette score, and therefore the framework can miss these nuances. 
Our framework does not limit the use of different clustering techniques so other methods of clustering can be explored. For example, replacing the clustering algorithm with another or using known initial cluster centres (eg: using HIGHLIGHTS \cite{amir2018highlights}) rather than random initializations could be explored.

Similarly for trajectory encoding, different independent sequence encoders such as LSTM or other transformer based encoders (eg: Vector Quantized Variational Auto-encoders) can be employed to extract the trajectory embeddings.

\section*{Acknolwedgement}

This work was supported in part by Taighde \'{E}ireann -- Research Ireland under grant numbers 13/RC/2077\_P2, 21/FFP-A/8957, and 18/CRT/6223. For the purpose of Open Access, the author has applied a CC BY public copyright license to any Author Accepted Manuscript version arising from this submission.

\bibliographystyle{splncs04} 
\bibliography{bibfile}

\newpage
\appendix

\section{Clustering}
\label{appendix_a}

\begin{figure}[!h]
     \centering
     \begin{subfigure}[b]{0.32\linewidth}
         \centering
         \includegraphics[width=\linewidth]{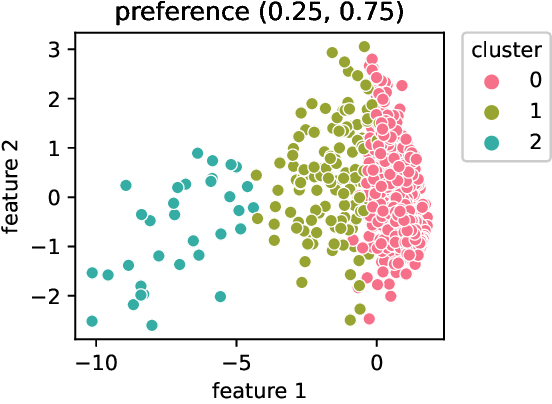}
         \caption{(0.25, 0.75)}
         \label{fig:a}
     \end{subfigure}
     \hfill 
     \begin{subfigure}[b]{0.32\linewidth}
         \centering
         \includegraphics[width=\linewidth]{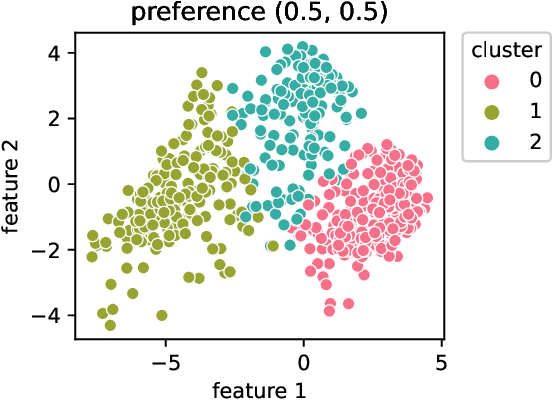}
         \caption{(0.5, 0.5)}
         \label{fig:b}
     \end{subfigure}
     \hfill
     \begin{subfigure}[b]{0.32\linewidth}
         \centering
         \includegraphics[width=\linewidth]{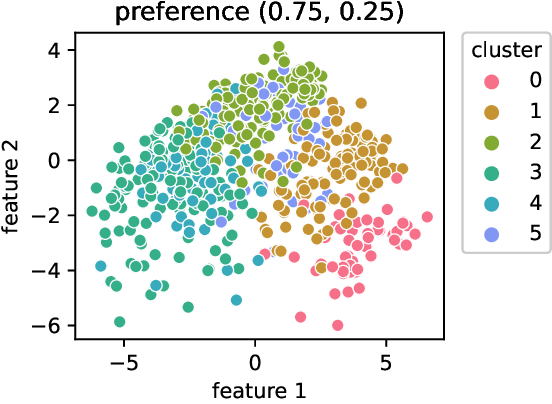}
         \caption{(0.75, 0.25)}
         \label{fig:c}
     \end{subfigure}
     \caption{MO-HalfCheetah}
     \label{fig:halfcheetah_clusters}
\end{figure}

\begin{figure}[!h]
     \centering
     \begin{subfigure}[b]{0.32\linewidth}
         \centering
         \includegraphics[width=\linewidth]{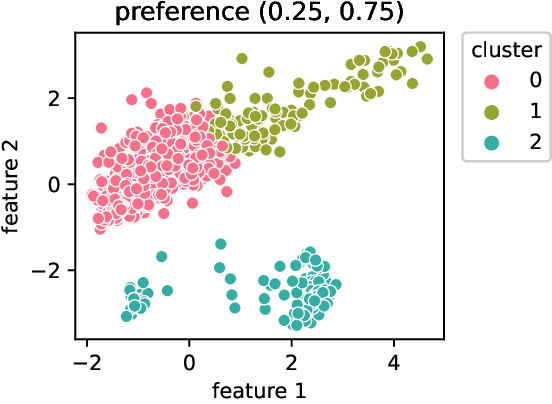}
         \caption{(0.25, 0.75)}
         \label{fig:ant_pref1}
     \end{subfigure}
     \hfill 
     \begin{subfigure}[b]{0.32\linewidth}
         \centering
         \includegraphics[width=\linewidth]{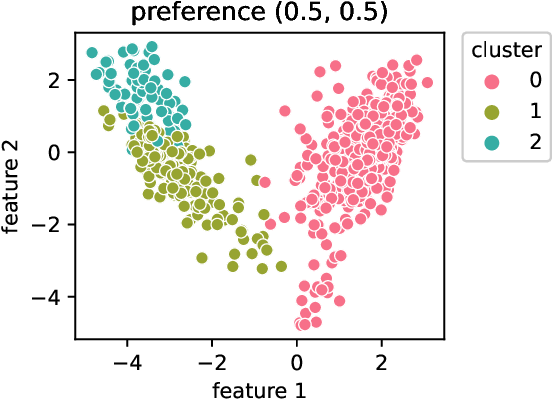}
         \caption{(0.5, 0.5)}
         \label{fig:b}
     \end{subfigure}
     \hfill
     \begin{subfigure}[b]{0.32\linewidth}
         \centering
         \includegraphics[width=\linewidth]{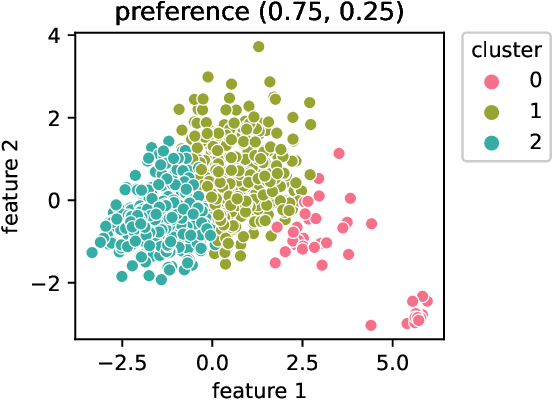}
         \caption{(0.75, 0.25)}
         \label{fig:c}
     \end{subfigure}
     \caption{MO-Ant}
     \label{fig:ant_clusters}
\end{figure}

\begin{figure}[!h]
     \centering
     \begin{subfigure}[b]{0.32\linewidth}
         \centering
         \includegraphics[width=\linewidth]{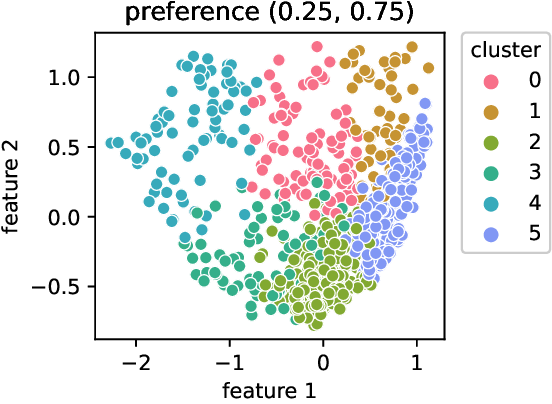}
         \caption{(0.25, 0.75)}
         \label{fig:a}
     \end{subfigure}
     \hfill 
     \begin{subfigure}[b]{0.32\linewidth}
         \centering
         \includegraphics[width=\linewidth]{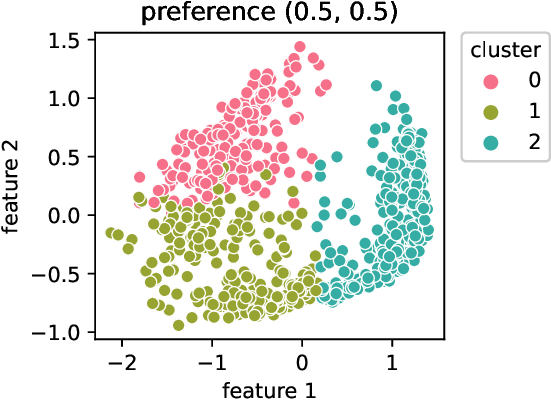}
         \caption{(0.5, 0.5)}
         \label{fig:b}
     \end{subfigure}
     \hfill
     \begin{subfigure}[b]{0.32\linewidth}
         \centering
         \includegraphics[width=\linewidth]{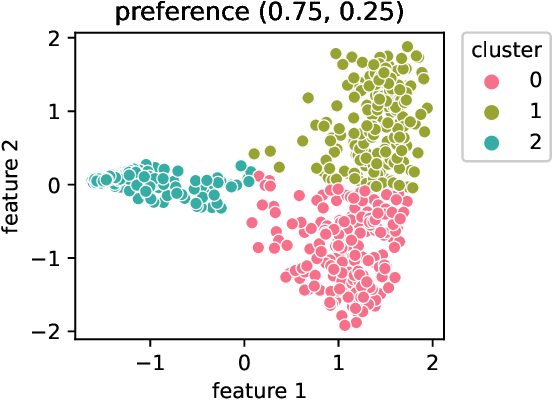}
         \caption{(0.75, 0.25)}
         \label{fig:c}
     \end{subfigure}
     \caption{MO-Swimmer}
     \label{fig:swimmer_clusters}
\end{figure}

In Sec.\ref{sec:preference_level}, we use K-means algorithm to cluster the sub-trajectories. The number of clusters are determined using the silhouette score for a range of k values. In Fig.\ref{fig:halfcheetah_clusters}--\ref{fig:swimmer_clusters} we illustrate the clusters for each environment MO-HalfCheetah, MO-Ant and MO-Swimmer under 3 preference settings - (0.25, 0,75), (0.5, 0,5) and (0.75, 0,25). Using Principle Component Analysis the high dimensional trajectory embeddings are reduced to 2-dimensions and shown in the figures. 

\vspace{-8mm}

\vspace{-8mm}

\end{document}